\pdfoutput=1

\documentclass[11pt]{article}

\usepackage[final]{acl}

\usepackage{amsmath}

\usepackage{times}
\usepackage{latexsym}

\usepackage{hyperref}
\usepackage{cleveref}

\usepackage{multirow}

\DeclareUnicodeCharacter{0268}{\ibar}

\usepackage[T1]{fontenc}

\usepackage[utf8]{inputenc}

\usepackage{microtype}

\usepackage{inconsolata}

\usepackage{graphicx}

\usepackage{tikz}

\usetikzlibrary{positioning, shapes.multipart}
\newcommand{\voc}[1]{\texttt{#1}}

\title{Can Uniform Meaning Representation Help GPT-4\\ Translate from Indigenous Languages?}

\author{Shira Wein \\
  Amherst College \\
  \texttt{swein@amherst.edu} }

\begin{document}
\maketitle
\begin{abstract}
While ChatGPT and GPT-based models are able to effectively perform many tasks without additional fine-tuning, they struggle with tasks related to extremely low-resource languages and indigenous languages. 
Uniform Meaning Representation (UMR), a semantic representation designed to capture the meaning of texts in many languages, is well-positioned to be leveraged in the development of low-resource language technologies.
In this work, we explore the downstream utility of UMR for low-resource languages by incorporating it into GPT-4 prompts.
Specifically, we examine the ability of GPT-4 to perform translation from three indigenous languages (Navajo, Arápaho, and Kukama), with and without demonstrations, as well as with and without UMR annotations. Ultimately, we find that in the majority of our test cases, integrating UMR into the prompt results in a statistically significant increase in performance, which is a promising indication of future applications of the UMR formalism.
\end{abstract}

\section{Introduction}

\begin{figure}[t]
\centering
\resizebox{\linewidth}{!}{
\begin{tikzpicture}[
blue/.style={rectangle, draw=black, very thick, minimum size=6mm},
]
	\node[blue](s) at (10,10) {\voc{buy-01}};
	\node[blue](p) at (7,7) {\voc{person}};
    \node[blue](t) at (6,3) {\voc{3rd}};
    \node[blue](sing1) at (8,3) {\voc{Plural}};
	\node[blue](a) at (9,7) {\voc{Activity}};
	\node[blue](f) at (11,7) {\voc{FullAff}};
    \node[blue](c) at (13,7) {\voc{car}};
    \node[blue](sing2) at (14,3) {\voc{Singular}};
    \node[blue](n) at (12,3) {\voc{new-01}};
	\draw[->, thick] (s.south) -- (p.north) node[midway, above, sloped] {\voc{:ARG0}};
	\draw[->, thick] (s.south) -- (c.north) node[midway, above, sloped] {\voc{:ARG1}};
 	\draw[->, thick] (s.south) -- (a.north) node[pos=0.7, sloped, above] {\voc{:aspect}};
 	\draw[->, thick] (s.south) -- (f.north) node[pos=0.65, sloped, above] {\voc{:modstr}};
    \draw[->, thick] (c.south) -- (sing2.north) node[midway, above, sloped] {\voc{:refer-number}};
    \draw[->, thick] (c.south) -- (n.north) node[midway, above, sloped] {\voc{:ARG1-of}};
    \draw[->, thick] (p.south) -- (sing1.north) node[midway, above, sloped] {\voc{:refer-number}};
    \draw[->, thick] (p.south) -- (t.north) node[midway, above, sloped] {\voc{:refer-person}};

\end{tikzpicture}}
\smallbreak
\small
\begin{verbatim}
(s / buy-01
  :ARG0 (p / person
    :refer-person 3rd
    :refer-number Plural)
  :ARG1 (c / car
    :ARG1-of (n / new-01)
    :refer-number Singular)
  :aspect Activity
  :modstr FullAff)
\end{verbatim}
\caption{UMR graph for the sentence ``They were buying a new car'' in both graph form and in text-based `PENMAN' notation \citep{kasper-1989-flexible}. 
}
\label{fig:umr_example}
\end{figure}


While ChatGPT models \citep{openai} are able to successfully produce text in many highly-resourced languages, they severely struggle with machine translation of low-resource languages
\citep{stap-araabi-2023-chatgpt,robinson-etal-2023-chatgpt}.

Uniform Meaning Representation \citep[UMR;][]{van2021designing} is a semantic representation created with the annotation of low-resource languages in mind.
The UMR formalism is designed to represent a wide range of languages by providing flexibility in the annotation process via paradigmatic lattices and creating any required rolesets during ``Stage 0'' of the annotation process.
UMR is a multilingual extension of the widely-adopted Abstract Meaning Representation \citep[AMR;][]{banarescu-etal-2013-abstract}.

The first UMR dataset \citep{bonn-etal-2024-building-broad} has recently been released, enabling exploration into the utility of UMR for tasks related to the generation of text into and from low-resource languages.
Recent work has also shown that GPT models likely do not implicitly contain the linguistic knowledge necessary to construct an AMR graph \citep{ettinger-etal-2023-expert}---or by extension, a UMR graph---suggesting that the addition of a UMR annotation may support prompt-based translation.

Thus, in this work, we explore the downstream benefits of incorporating UMR graphs into ChatGPT prompts, specifically with regard to machine translation from extremely low-resource languages into English. 
We craft four prompting protocols for GPT-4\footnote{\url{https://openai.com/index/gpt-4/}} which vary in both their number of demonstrations and whether UMR is included: (1)~zero-shot prompting, (2)~zero-shot prompting with the UMR graph of the text included, (3)~five-shot prompting, and (4)~five-shot prompting with the UMR graphs included.
We perform our experiments on three indigenous languages included in \citet{bonn-etal-2024-building-broad} which also contain English references: Navajo, Kukama, and Arápaho.
Our contributions include:
\begin{itemize}
\addtolength\itemsep{-2mm}
    \item Prompting protocols for translating from indigenous languages, with and without demonstrations (i.e. zero- and five-shot), and with and without UMR graphs of the source text. 
    \item Experiments producing English translations of more than 1000 individual source sentences across three extremely-low resource languages via GPT-4.
    \item Statistical analyses of the results of each of our protocols, which indicate the quantitative improvement that the incorporation of UMR graphs and demonstrations begets.
\end{itemize}

\section{Background \& Related Work}

\subsection{Machine Translation with ChatGPT}

Recent work has explored the effectiveness of prompting GPT models to generate text in no- and low-resource languages, with generally poor indications of success \citep{stap-araabi-2023-chatgpt}. 

\citet{guo-etal-2024-teaching} focus on mitigating the issue of data sparsity for low-resource translation via ChatGPT and BLOOMZ \citep{muennighoff-etal-2023-crosslingual} by providing a vocabulary list and demonstrations as additional input.

Notably, \citet{robinson-etal-2023-chatgpt} find that ChatGPT performs competitively with state-of-the-art machine translation models for high-resource languages, but performs poorly for low-resource languages. In particular, the most significant predictor of ChatGPT translation performance on a language is the number of Wikipedia entries that exist in the language, serving as a proxy of how well-sourced that language is.
Additionally, five-shot prompts lead to small performance gains over zero-shot prompts.
\citet{Tang_Qin_Ye_Tan_Yang_2025} further indicate that (in a high-resource setting) adaptive few-shot prompting, which uses the most semantically similar texts in the dataset to the source text as demonstrations, leads to increased performance gains.

Related work has explored the utility of chain-of-thought prompting for translating with ChatGPT, finding it to be generally ineffective as it results in word-by-word translation \citep{peng-etal-2023-towards}.

\subsection{Uniform Meaning Representation}

Uniform Meaning Representation (UMR) is an extension of the popular semantic representation Abstract Meaning Representation (AMR).
AMR is a graph-based semantic representation which captures ``who does what to whom,'' reflecting the semantic relationships within the sentence.
The nodes in an AMR graph correspond with concepts in the sentence (or phrase), while edges denote the relationships between those concepts. 
UMR, like AMR, represents the relationships between concepts in a sentence in the form of a rooted, directed graph (see \Cref{fig:umr_example} for an example UMR in both text-based and graph-based form).

While AMR was originally designed for English \citep{banarescu-etal-2013-abstract}, UMR is designed to be multilingual and contains information about the text at both the sentence- and document-level \citep{van2021designing}.
UMR accommodates a range of linguistic features in comparison to AMR \citep{wein-bonn-2023-comparing} through the integration of lattice-based annotation structures, which allow the annotator to select the level of granularity appropriate for the individual language  \citep{van-gysel-etal-2019-cross}.
UMR is also particularly well-suited to the annotation of low- and no-resource languages, including indigenous languages \citep{van-gysel-etal-2021-theoretical}, as it incorporates dataset development (in the form of rolesets) into ``Stage 0'' of the annotation process \citep{vigus-etal-2020-cross}, thus overcoming the lack of preexisting rolesets for some languages. 
UMR's design, which both accommodates linguistic diversity and overcomes a lack of data for low-resource languages, motivates its use in our work.


While AMR is a semantic representation that has been widely adopted and has proven useful in many monolingual settings \citep{wein-opitz-2024-survey}, it is not possible to annotate low-resource languages in AMR.
A language-specific version of AMR would need to be created for each individual low-resource language, as there is not sufficient annotation flexibility in AMR to effectively annotate multilingually \citep{wein-schneider-2024-assessing}, thus making AMR not appropriate for our work.
Still, prior work has demonstrated that AMR is particularly useful in low-resource engineering studies \citep{hua-etal-2023-improving,gururaja-etal-2023-linguistic,ghosh-etal-2024-abex}, which motivates incorporating UMR into low-resource settings and for low-resource languages.

\section{Methodology}
\subsection{Data}
In this work, we examine the utility of incorporating UMR graphs into GPT-4 prompts which instruct the system to translate a source text in the extremely low-resource languages of Navajo, Kukama, and Arápaho into English.

The recently released UMR dataset which is leveraged in this work \citep{bonn-etal-2024-building-broad} contains sentences from English (209 sentence-level graphs, 202 document-level), Chinese (358 sentence-level graphs, 358 document-level), Arápaho (406 sentence-level graphs, 109 document-level), Navajo (506 sentence-level graphs, 168 document-level), Kukama (105 sentence-level graphs, 86 document-level), and Sanapaná (602 sentence-level graphs, 602 document-level).
Arápaho, Navajo, Kukama, and Sanapaná are all indigenous languages which are extremely low-resource.
Not all annotations contain both sentence-level and document-level graphs.
The Navajo, Kukama, and Arápaho UMR graphs all provided English translations with the annotations, while the Sanapaná annotations contained Spanish translations.
While included in the UMR dataset, we forgo translation from Sanapaná, as English translations are not provided, negating our ability to use English texts as references when evaluating the system output.

We generate translations from the 506 sentences in Navajo, 105 sentences in Kukama, and 406 sentences in Arápaho, resulting in 1,017 total sentences translated.

\subsection{Prompts}
We design and use four prompting protocols:
\begin{enumerate}
\addtolength\itemsep{-2mm}
    \item Zero-shot: Instruct the model to translate from the source language into English, providing the text to be translated.
    \item Zero-shot with UMR: Instruct the model to translate from the source language into English, providing the text to be translated and the UMR of the source text.
    \item Five-shot: Instruct the model to translate from the source language into English, providing five demonstrations (texts in the source language as well as their reference English translations), as well as the text to be translated.
    \item Five-shot with UMR: Instruct the model to translate from the source language into English, providing five demonstrations (texts in the source language as well as their reference English translations, plus their UMRs), the text to be translated, and the UMR of the source text.
\end{enumerate}

For our five-shot prompts, we use an adaptive approach to demonstration selection, selecting the 5-nearest neighbors to the source sentence as the demonstrations. 
We use chrF to compare the source language text to the other sentences in that language.
We are using the source language sentences (rather than the English references) to identify the five most relevant demonstrations, to ensure that the same would be possible at test time.

The specific text contained in each prompt can be seen in \Cref{fig:prompts} in \Cref{sec:appendix}.

\subsection{Evaluation}
We perform translation \emph{from} the indigenous languages \emph{into} English in order to enable more accurate evaluation of the generated text, via automatic and qualitative analyses.

We evaluate the performance of the model for each item using each of the four prompting protocols.\footnote{This experimentation cost \$62.11 USD in OpenAI credits.}
The automatic metrics we use to evaluate our generated text are  chrF \citep{popovic-2015-chrf} and BERTscore \citep{zhang2019bertscore}.

\section{Results}
\label{sec:results}

\begin{table*}[tb]
    \centering
    \small
    \begin{tabular}{|c|c|c|c|c|}
    \hline
    Evaluation Metric & Prompting Protocol & Arápaho & Kukama & Navajo \\
    \hline
    \multirow{4}{5em}{BERTscore} & Zero-Shot & 0.867$\pm$0.02 & 0.862$\pm$0.02 & 0.862$\pm$0.02 \\
    & Zero-Shot w UMR & 0.867$\pm$0.05 & 0.857$\pm$0.03 & 0.867$\pm$0.03 \\
    & Five-Shot & 0.903$\pm$0.04 & 0.904$\pm$0.04 & 0.885$\pm$0.03 \\
    & Five-Shot w UMR & 0.910$\pm$0.04 & 0.912$\pm$0.04 & 0.891$\pm$0.03 \\
    \hline
    \multirow{4}{2em}{chrF} & Zero-Shot & 13.0$\pm$5.5 & 14.0$\pm$5.8 & 15.4$\pm$6.4 \\
    & Zero-Shot w UMR & 16.2$\pm$8.7 & 16.8$\pm$7.0 & 17.9$\pm$8.3 \\
    & Five-Shot & 32.9$\pm$21 & 40.8$\pm$25 & 24.6$\pm$14.2 \\
    & Five-Shot w UMR & 35.7$\pm$22 & 43.5$\pm$24 & 25.9$\pm$14.1 \\
    \hline
    \end{tabular}
    \caption{Average scores for each language, prompting protocol, and evaluation metric. Standard deviation is indicated after the plus or minus sign.}
    \label{tab:avg_bertscore}
\end{table*}


\begin{table*}[tb]
    \centering
    \small
    \begin{tabular}{|c|c|c|c|c|}
    \hline
    & Arápaho & Kukama & Navajo \\
    \hline
    BERTScore: Zero-shot vs Zero-shot with UMR & 0.9721 & \textcolor{red}{0.0146} & \textbf{<0.0001} \\
    chrF: Zero-shot vs Zero-shot with UMR & \textbf{<0.0001} & \textbf{<0.0001} & \textbf{<0.0001} \\
    \hline
    BERTScore: Zero-shot vs Five-shot & \textbf{<0.0001} & \textbf{<0.0001} & \textbf{<0.0001} \\
    chrF: Zero-shot vs Five-shot & \textbf{<0.0001} & \textbf{<0.0001} &  \textbf{<0.0001}\\
    \hline
    BERTScore: Five-shot vs Five-shot with UMR & \textbf{<0.0001} & \textbf{0.0017} & \textbf{<0.0001} \\
    chrF: Five-shot vs Five-shot with UMR & \textbf{0.0004} & 0.0555 & \textbf{0.0294} \\
    \hline
    \end{tabular}
    \caption{Two-tailed paired t-test p-values for statistical comparisons of the BERTscores and chrF scores for each prompting protocol, in each language. The bolded entries indicate a statistically significant improvement when adding UMR or demonstrations. The red entry (first row of Kukama scores) highlights a statistically significant difference, but where the zero-shot \emph{without} UMR performs better than the zero-shot \emph{with} UMR.}
    \label{tab:p-values}
\end{table*}

\Cref{tab:avg_bertscore} shows the average BERTscore and chrF values, respectively, of each prompting protocol in each language.
While the BERTscore values are all closer together (as expected, given that BERTscore struggles to capture finer-grained differences in meaning \citep{leung-etal-2022-semantic,wein-etal-2023-measuring}), we can see generally that for BERTscore, the five-shot with UMR scores are highest, followed by five-shot scores.
Then, with a notable decrease in BERTscore value, the zero-shot and zero-shot with UMR scores follow, but switching which of the two is higher.
For the chrF scores, the five-shot with UMR scores are highest for all languages, followed by the five-shot scores, next followed by the zero-shot with UMR scores (again notably lower than the five-shot performance) and finally the zero-shot scores.

While these average scores provide an initial impression as to the benefits of adding UMR graphs and demonstrations in the prompt, we perform two-tailed paired t-tests comparing the automatic metric scores of the output from the various prompts
in order to ascertain whether the inclusion of a UMR graph and\slash or the five demonstrations results in a statistically significant increase in the score (and therefore, performance).
Specifically, we compare the scores for (1)~zero-shot and zero-shot with UMR prompts, (2)~zero-shot and five-shot prompts, and (3)~five-shot and five-shot with UMR prompts. These values can be seen in \Cref{tab:p-values}. 

We find that for 9 of the 12 comparisons of prompts with UMR versus without UMR, adding the UMR to the prompt results in a statistically significant increase. 
For two comparisons (Arápaho BERTscore: Zero-shot vs Zero-shot with UMR; Kukama chrF: Five-shot versus Five-shot with UMR), there is no statistical difference, and in one case (Kukama BERTscore zero-shot vs zeros-show with UMR), there is a statistically significant difference but the scores from the prompt without UMR are higher.
These findings indicate that the inclusion of a UMR graph into a prompt enables more effective text generation in extremely low-resource settings, as the UMR may be supplying additional linguistic information not already contained in the model.

Additionally, for all 6 of the cases where we compare zero-shot scores against five-shot scores, there is a statistically significant improvement.
This indicates that the use of demonstrations is also useful for prompting with extremely low-resource languages.
While both UMR and demonstrations lead to improvements in translation quality, the most drastic difference is found when adding the demonstrations to move from zero-shot to five-shot prompting.

As indicated by our quantitative results and statistical analysis, our qualitative analysis further suggests that the English translation more closely resembles the reference when incorporating UMR and\slash or demonstrations into the prompt. 

Take as an example the following Kukama item, which has a disfluent English reference of ``He run in the forest:''
\includegraphics[trim={2cm 13.4cm 2.2cm 13.3cm},clip,scale=0.24]{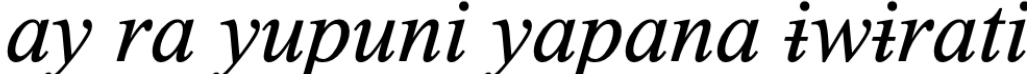}.
The zero-shot translation is completely unrelated to the text, though it does have some reference to a man performing an action in a natural setting: ``He plays with his younger brother at the river.''
The zero-shot with UMR text is again unrelated, indicating a person performing an action in an unspecified setting: ``The person is working there today.''
Then, the five-shot text contains more semantic similarity with the reference, as there is a male moving in the forest: ``He has already started walking in the forest.''
Finally, the five-shot text with UMR shows the male to be running in the forest, clearly exhibiting the most semantic similarity with the reference: ``He has already started running in the forest.''
This example exemplifies the fact that five-shot prompting alone is not enough to achieve optimal performance on this task, and leveraging UMR in the prompt benefits performance beyond what is possible only when using demonstrations. 

Another example is the Arápaho sentence, \emph{wohei noh ci'ceese' hoo3itoo, heetnoo3itoone3en}, which has an English reference of ``Wohei another story, I'm going to tell you another story.''
The zero-shot text is completely unrelated to the reference: ``I walked to the store and said hello to the shopkeeper.''
The zero-shot with UMR output, on the other hand, is much more semantically similar to the reference: ``I will tell you a story.''
The five-shot text then contains some of the specific language included in the reference, as well as some semantic similarity: ``Wohei, then go ask him, the storyteller.''
Finally, the five-shot with UMR text is the most semantically similar, though the ending is slightly harder to interpret: ``Wohei I will tell you a story about a little.'' 

Our automatic metrics and qualitative analyses reveal that, on our test data, for most cases, incorporation of UMR graphs and demonstrations into the prompts enables heightened similarity with the reference English translation.
Therefore, leveraging UMR in the prompt does indeed lead to heightened performance when translating from indigenous languages into English, while the greatest improvements are achieved by using related sentences as demonstrations; the combination of using both the demonstrations and the UMR in the prompt leads to the highest quality output in our experiments.

\section{Conclusion}
In this work, we begin to address one of the failings of GPT-based models: that of translation from extremely low-resource languages. We specifically examine the ability of a newly released Uniform Meaning Representation (UMR) dataset---containing sentences in Navajo, Arápaho, and Kukama, their UMR graphs, and their parallel sentences in English---to improve GPT-4 performance when included in the prompt.
We find that both the incorporation of UMR graphs of the source text and adaptively selected demonstrations lead to improved performance on low-resource machine translation via prompting, with a statistically significant increase resulting in the majority of our comparisons.
This is a promising indication of the downstream utility of UMR for low-resource settings and a step forward towards effective translation from indigenous languages via prompting.

\section*{Limitations}

We perform experimentation on three extremely low-resource indigenous languages.
Future work could expand this evaluation to other languages as well, varying in their depth of resources, as additional UMR annotations are released.
UMR annotation can be expensive and time-consuming, as it requires fluency in the language and annotator training, which is a barrier to seamlessly incorporating UMR into downstream applications.

Additionally, we perform translation in one direction, with the low-resource languages serving as the source and English serving as the target language. Performing translation from English into these low-resource languages would make for interesting future work, though it will require a human evaluation by speakers of the language.

Finally, randomness is inherent in the results generated from GPT-4. We attempt to curtail this effect by providing statistical analyses for our findings, but further rigor could be added by running these experiments additional times.


\bibliography{custom}

\appendix

\section{Prompting Protocols}
\label{sec:appendix}

\begin{figure*}[tbp]
\begin{tabular}{l}
\small
\emph{Zero-shot}\\
\small\texttt{Please provide the English translation for this [Source language] sentence. Do not provide}\\
\small \texttt{any explanations or text apart from the translation.} \\
\small \texttt{[Source Language]: [sentence to be translated]} \\
\small \texttt{English:}\\
\hline
\small\emph{Zero-shot with UMR}\\ 
\small\texttt{Please provide the English translation for this [Source language] sentence (which is} \\
\small\texttt{accompanied by a Uniform Meaning Representation parse). Do not provide any explanations or text apart} \\
\small\texttt{from the translation.}\\
\small\texttt{[Source language]: [sentence to be translated]}\\
\small\texttt{Uniform Meaning Representation: [UMR of source text]}\\
\small\texttt{English:} \\
\hline
\small\emph{Five-shot}\\
\small\texttt{Please provide the English translation for this [Source language] sentence. Do not provide}\\
\small\texttt{any explanations or text apart from the translation.} \\
\small\texttt{[Source language]: [sentence 1] English: [translation 1]} \\
\small\texttt{[Source language]: [sentence 2] English: [translation 2]} \\
\small\texttt{[Source language]: [sentence 3] English: [translation 3]} \\
\small\texttt{[Source language]: [sentence 4] English: [translation 4]} \\
\small\texttt{[Source language]: [sentence 5] English: [translation 5]} \\
\small\texttt{Please provide the English translation for this [Source language] sentence.} \\
\small\texttt{Do not provide any explanations or text apart from the translation.} \\
\small\texttt{[Source language]: [sentence to be translated]} \\
\small\texttt{English:} \\
\hline
\small\emph{Five-shot with UMR}\\
\small\texttt{Please provide the English translation for this [Source language] sentence (which is} \\
\small\texttt{accompanied by a Uniform Meaning Representation parse). Do not provide any explanations or text apart}\\
\small\texttt{from the translation.} \\
\small\texttt{[Source language]: [sentence 1] Uniform Meaning Representation: [UMR 1] English: [translation 1]} \\
\small\texttt{[Source language]: [sentence 2] Uniform Meaning Representation: [UMR 2] English: [translation 2]} \\
\small\texttt{[Source language]: [sentence 3] Uniform Meaning Representation: [UMR 3] English: [translation 3]} \\
\small\texttt{[Source language]: [sentence 4] Uniform Meaning Representation: [UMR 4] English: [translation 4]} \\
\small\texttt{[Source language]: [sentence 5] Uniform Meaning Representation: [UMR 5] English: [translation 5]} \\
\small\texttt{Please provide the English translation for this [Source language] sentence (which is accompanied by}\\
\small\texttt{a Uniform Meaning Representation parse).
Do not provide any explanations or text apart from the}\\
\small\texttt{translation.} \\
\small\texttt{[Source language]: [sentence to be translated]} \\
\small\texttt{Uniform Meaning Representation: [UMR of source text]} \\
\small\texttt{English:} \\
\end{tabular}
\caption{The ``user'' portions of our prompts for the four protocols. For all prompts, the ``system'' portion of the protocol is as follows:
\texttt{System: You are a machine translation system from [Source language] to English that translates sentences from narrative documents.}
}
\label{fig:prompts}
\end{figure*}

\end{document}